\pdfoutput=1

\documentclass[11pt]{article}

\usepackage[final]{acl}

\usepackage{times}
\usepackage{latexsym}

\usepackage[T1]{fontenc}

\usepackage[utf8]{inputenc}

\usepackage{microtype}

\usepackage{inconsolata}

\usepackage{natbib}
\usepackage{multirow}
\usepackage{booktabs}
\usepackage{graphicx}

\title{Fine-tuning Large Language Models for Multigenerator, Multidomain, and Multilingual Machine-Generated Text Detection}

\author{Feng Xiong\textsuperscript{1} \and Thanet Markchom\textsuperscript{2} \and Ziwei Zheng\textsuperscript{1} \and \\ \textbf{Subin Jung}\textsuperscript{1} \and \textbf{Varun Ojha}\textsuperscript{1} \and \textbf{Huizhi Liang}\textsuperscript{1} \\
         \textsuperscript{1}School of Computing, Newcastle University, Newcastle upon Tyne, UK \\
         \textsuperscript{2}Department of Computer Science, University of Reading, Reading, UK \\ 
         \texttt{xf199912@163.com, t.markchom@pgr.reading.ac.uk, \{z.zheng21,}\\
         \texttt{s.jung4, varun.ojha, huizhi.liang\}@newcastle.ac.uk}}

\begin{document}
\maketitle
\begin{abstract}

SemEval-2024 Task 8 introduces the challenge of identifying machine-generated texts from diverse Large Language Models (LLMs) in various languages and domains. The task comprises three subtasks: binary classification in monolingual and multilingual (Subtask A), multi-class classification (Subtask B), and mixed text detection (Subtask C). This paper focuses on Subtask A \& B. Each subtask is supported by three datasets for training, development, and testing. To tackle this task, two methods: 1) using traditional machine learning (ML) with natural language preprocessing (NLP) for feature extraction, and 2) fine-tuning LLMs for text classification. The results show that transformer models, particularly LoRA-RoBERTa, exceed traditional ML methods in effectiveness, with majority voting being particularly effective in multilingual contexts for identifying machine-generated texts.

\end{abstract}

\section{Introduction}

Large Language Models (LLMs) are sophisticated natural language processing (NLP) models extensively trained on vast textual datasets~\cite{wang2023m4}. These models demonstrate an impressive proficiency in generating human-like text based on the input they receive.  Whether completing sentences, generating paragraphs, or even crafting entire articles, LLMs showcase unparalleled proficiency in mimicking natural languages~\cite{mitchell2023detectgpt}. However, using LLMs for generating texts has raised concerns about potential misuse, such as disseminating misinformation and disruptions in the education system~\cite{wang2023m4}. Therefore, there is an urgent need to develop automated systems for detecting machine-generated texts~\cite{mitchell2023detectgpt, wang2023m4}.

Recently, several LLMs have been developed such as ChatGPT\footnote{https://chat.openai.com/}~\cite{brown2020language}, Cohere\footnote{https://cohere.com}, Davinci\footnote{https://platform.openai.com/docs/models/gpt-base}, BLOOMZ\footnote{https://huggingface.co/bigscience/bloomz}~\cite{muennighoff2022crosslingual}, and Dolly\footnote{https://huggingface.co/databricks/dolly-v2-12b}~\cite{DatabricksBlog2023DollyV2}.
The versatility of these models extends across various domains, such as news, social media, educational platforms, and academic contexts, in multiple languages not only English~\cite{wang2023m4}. This wide application poses a challenge in developing an automated system capable of detecting machine-generated texts from various generators, across multiple domains and languages. 

To tackle this challenge, SemEval-2024 Task 8: Multigenerator, Multidomain, and Multilingual Black-Box Machine-Generated Text Detection~\cite{wang2023m4} introduces the task of detecting machine-generated texts obtained from different LLMs, in various domains and languages. This task consists of three subtasks: Subtasks A, B, and C. Subtask A involves binary classification of text as either human-written or machine-generated, with two tracks: monolingual (English only) and multilingual.
Subtask B focuses on multi-class classification of machine-generated text, aiming to identify the source of generation, whether human or a specific language model.
Subtask C addresses the detection of human-machine mixed text, requiring the determination of the boundary where the transition from human-written to machine-generated occurs in a mixed text. This paper focuses on Subtasks A and B. 
To tackle these tasks, we propose two approaches: (1) classical machine learning, leveraging NLP techniques for feature extraction, and (2) fine-tuning LLMs for the classification of human-written and machine-generated texts.

\section{Related Work}
In the realm of distinguishing texts generated by large language models (LLMs) from those authored by humans, researchers have employed a variety of methods and tools. Broadly, these approaches can be categorized into two main types: black-box and white-box detection methods \cite{tang2023science}. Black-box detection relies on API-level access to LLMs, utilizing textual samples from both human and machine sources to train classification models \cite{dugan2020roft}. The study by Guo et al. (2023) integrated existing question-and-answer datasets and leveraged fine-tuning of pre-trained models to investigate the characteristics and similarities between human-generated and AI-generated texts \cite{guo2023close}.

In the context of white-box detection, Kirchenbauer et al. (2023) introduced a novel approach involving the embedding of watermarks in the outputs of large language models to facilitate the detection of AI-generated text \cite{kirchenbauer2023watermark}. Additionally, a variety of tools and methodologies, including XGBoost, decision trees, and transformer-based models, have been evaluated for their efficacy in detecting texts produced by AI \cite{zaitsu2023distinguishing}. These techniques incorporate multiple stylistic measurement features to differentiate between AI-generated and human-generated texts \cite{shijaku2023chatgpt}.

Specific tools and techniques in this domain include the GLTR tool developed by Gehrmann et al. (2019), which analyzes the usage of rare words in texts to distinguish between those generated by the GPT-2 model and human writers \cite{gehrmann2019gltr}. The DetectGPT method posits that minor rewrites of LLM-generated texts tend to reduce the log probability under the model, a hypothesis that has been explored in depth \cite{mitchell2023detectgpt}. Furthermore, intrinsic dimension analysis, including methods like the Persistent Homology Dimension estimator (PHD), has been applied to distinguish between authentic texts and those generated artificially \cite{tulchinskii2023intrinsic}.

Detectors specifically designed for certain LLMs, such as the GROVER detector for the GROVER model \cite{zellers2019defending} and the RoBERTa detector using the RoBERTa model \cite{liu2019roberta}, also play a significant role in this field. The development of these detectors reflects the ongoing exploration and refinement in the field to discern the increasingly sophisticated outputs of LLMs from human writing.

In summary, the combination of statistical analysis with advanced language models is being employed by researchers to more effectively differentiate between content generated by humans and machines. The continuous evolution and refinement of these techniques reflect the dynamic nature of the field and the complexities involved in distinguishing between the increasingly nuanced outputs of LLMs and human-authored texts.

\section{Methods}

To tackle these tasks, we employ two distinct strategies. The first is classical machine learning, tailored for natural language preprocessing (NLP). The second approach involves transformer-based LLMs, with an emphasis on LoRA (Low-Rank Adaptation of Large Language Models) fine-tuning~\cite{hu2021lora}. We then enhance our results by integrating these methods through ensemble techniques.

\subsection{Machine Learning Models}

Our approach for textual data analysis in machine learning involves a concise yet comprehensive preprocessing pipeline. Initially, URLs and excess whitespace are removed from the text. Next, all punctuation is eliminated, focusing solely on alphanumeric characters. The text is further refined by excluding common stopwords and numeric characters. Emojis are decoded into text, providing additional context. Lemmatization standardizes words to their base forms, ensuring consistent analysis. Texts are then converted to lowercase for uniformity. The final step involves using a \emph{Term Frequency-Inverse Document Frequency} (TF-IDF), configured to handle a maximum of 8000 features and considering unigrams to trigrams. This vectorizer excludes terms appearing in less than 10 documents, balancing feature representation with computational efficiency. 

To enhance the feature set for machine learning, we incorporate the \emph{Gunning fog index}~\cite{readabilitygunning} and \emph{Flesch reading ease score}~\cite{kincaid1975derivation} into our text analysis. The Gunning Fog Index is a readability metric that estimates the years of formal education needed to understand a text on the first reading. It's calculated based on sentence length and the complexity of words used. A higher index indicates more complex text. The Flesch reading ease score, on the other hand, assesses the readability of a text based on sentence length and syllable count per word. Scores range from 0 to 100, with higher scores indicating easier readability.

This preprocessing strategy transforms raw text into a structured numerical format, ready for machine learning model analysis. 

In our study, we employed three distinct machine learning algorithms for both binary and multi-class classification tasks: Logistic Regression (LR), Multinomial Naive Bayes Classifier (MultinomialNB), and eXtreme Gradient Boosting (XGBoost)~\cite{chen2016xgboost}.

\begin{itemize}
    \item LR: A linear model used for classification tasks. It models the probability that a given input belongs to a certain class. Logistic Regression is particularly effective for binary classification due to its simplicity and efficiency in estimating probabilities.

    \item MultinomialNB: This algorithm is based on the Bayes theorem and is particularly suited for classification with discrete features (like word counts for text classification). It assumes independence between predictors and is highly scalable to large datasets.

    \item XGBoost: This is an efficient and scalable implementation of gradient-boosted decision trees. It is known for its performance and speed, especially in structured or tabular data, and can handle both binary and multi-class classification problems effectively.
\end{itemize}

By integrating these algorithms, our approach leverages the strengths of linear modeling, probabilistic classification, and ensemble learning, aiming to enhance predictive accuracy and robustness across diverse classification scenarios.

\subsection{XLM-RoBERTa}

In our approach, we established XLM-RoBERTa\footnote{https://huggingface.co/xlm-roberta-base}~\cite{conneau2019unsupervised} as the baseline model among transformer-based architectures. XLM-RoBERTa represents a multilingual adaptation of the original RoBERTa~\cite{liu2019roberta} model, specifically designed to understand and process a diverse range of languages. XLM-RoBERTa is pre-trained on a substantial dataset: 2.5TB of filtered CommonCrawl data~\cite{zhang2020semi}, encompassing text in 100 different languages. This extensive pre-training enables the model to capture nuanced language features and patterns across a broad linguistic spectrum, making it highly effective for tasks involving multiple languages. The use of such a diverse training dataset aids in achieving a robust understanding of various linguistic structures and vocabularies, which is crucial for accurate language processing and analysis in a multilingual context.

\subsection{LoRA-RoBERTa}

\begin{figure}[htb!]
    \centering
    \includegraphics[width=0.3\textwidth]{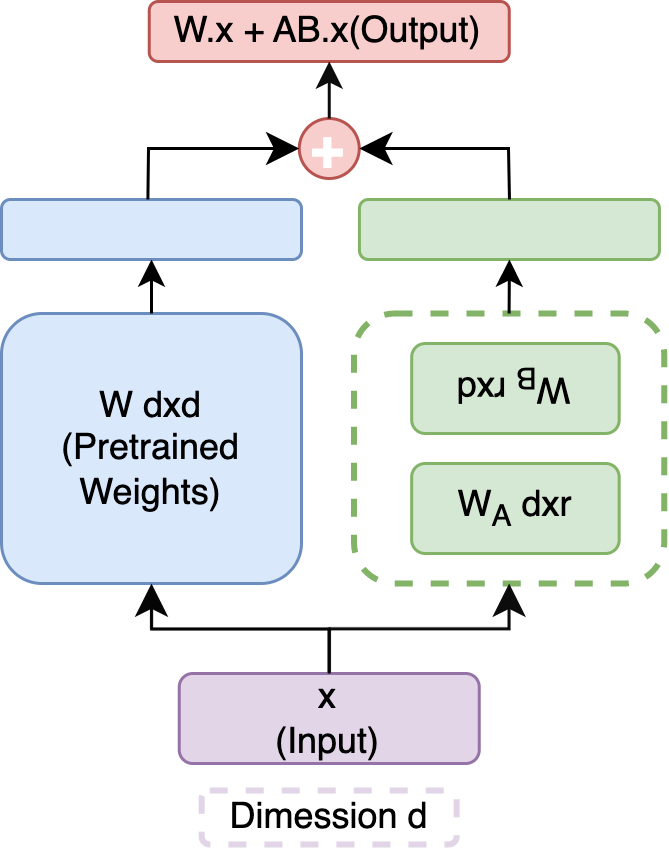}
    \caption{LoRA-based fine-tuning streamlines the process by freezing the original weights of LLMs and training a minimal number of parameters. This approach significantly reduces resource consumption compared to classic fine-tuning, which alters most model weights}
    \label{fig:1}
\end{figure}

To further improve the predictive performance of LLMs, we use LoRA for fine-tuning RoBERTa\footnote{https://huggingface.co/roberta-base} model. LoRA is a technique enhancing the efficiency of fine-tuning large models with reduced memory consumption. It modifies the weight updates in neural networks using two smaller matrices derived through low-rank decomposition. These matrices adapt to new data while the original weights remain unchanged. The final output combines the original and adapted weights. In transformer models, LoRA is often applied to attention blocks for efficiency. The number of trainable parameters depends on the low-rank matrices' size, influenced by the rank and the original weight matrix's shape~\cite{hu2021lora}, as shown in Figure \ref{fig:1}.

\subsection{Majority Voting}

Majority voting in ensemble learning, where the final decision is based on the majority vote from multiple machine learning models, offers several advantages over a single-model approach. This technique, applicable in scenarios with 
$N$ classifiers ($C_1, C_2,\ldots,C_N$), determines the final output $V(x)$ as the class receiving the most votes: $V(x) = \text{mode}\{ C_1(x), C_2(x), \ldots, C_N(x) \}$. This method effectively reduces variance by balancing out individual model errors, leading to more stable predictions. Furthermore, it generally achieves higher accuracy due to the diverse perspectives of different models. Its robustness against overfitting is enhanced, as it combines various models' strengths, making it suitable for a wider range of data scenarios. The flexibility in model selection allows for a blend of different algorithms, each capturing unique data patterns, which contributes to better generalization on unseen data. Thus, majority voting stands out as a robust, accurate, and flexible approach in machine learning.

\subsection{DistilmBERT}

RoBERTa and XLM-RoBERTa are both powerful but computationally expensive models. Therefore, we investigate an alternative model that is more computationally efficient, aiming to compare its performance against these models. We adopted \emph{DistilBERT base multilingual cased}\footnote{https://huggingface.co/distilbert-base-multilingual-cased} (DistilmBERT)~\cite{Sanh2019DistilBERTAD} which is a distilled version of the BERT base multilingual model, pretrained on the concatenation of Wikipedia in 104 different languages. DistilmBERT consists of 6 layers, each with 768 dimensions and 12 attention heads, totaling 134 million parameters. This configuration balances model efficiency while retaining significant representational power~\cite{Sanh2019DistilBERTAD}.

\section{Experiments}

In our study, subtask A focuses on distinguishing between human-written (label 0) and machine-generated text (label 1), offered in both monolingual (119,757 train, 5,000 dev, 34,272 test) and multilingual versions (172,417 train, 4,000 dev, 42,378 test), across various sources and languages are given in Table \ref{tab:data}. Subtask B, with 71,027 train, 3,000 dev, and 18,000 test, goes further by identifying the specific model (including ChatGPT, Cohere, DaVinci, BloomZ, and Dolly) that generated the text, or if it's human-generated. Both tasks utilize datasets with an identifier, label, text content, model name, and source, focusing on the nuanced classification of texts.

\begin{table}[htb!]
\centering
\begin{tabular}{llll}
\hline
Subtask          & \#Train & \#Dev & \#Test \\ \hline
A - Monolingual  & 119,757 & 5,000 & 34,272 \\
A - Multilingual & 172,417 & 4,000 & 42,378 \\
B                & 71,027  & 3,000 & 18,000 \\ \hline
\end{tabular}
\caption{Dataset for text classification subtasks}
\label{tab:data}
\end{table}

\subsection{Parameter Settings}

In our experimentation, hyperparameter settings varied between classical machine learning models and LLMs.

For the classical machine learning models, we adhered to default parameter settings during training. This approach simplifies the process and relies on the general applicability of these preset parameters.

In contrast, for LLMs, specific hyperparameters were carefully chosen. When training the XLM-RoBERTa baseline model, we set the batch size to 16 and the learning rate to $2.0e-5$ with the model being trained for 3 epochs. This configuration ensures efficient handling of data and optimal learning speed. For fine-tuning the LoRA-RoBERTa base model, the learning rate was adjusted to $1.0e-3$ over 5 epochs, a setting conducive to the specific demands of fine-tuning.

Furthermore, we employed configuration for the LoRA fine-tuning, defined with the following parameters: \emph{task\_type} set to \emph{SEQ\_CLS} indicating a sequence classification task, \emph{r} (rank of the low-rank matrices) set to 4, \emph{lora\_alpha} (scaling factor for learning rate) at 32, \emph{lora\_dropout} to manage overfitting set at 0.01, and \emph{target\_modules} focused on the \emph{query} module. These configurations are critical in guiding the fine-tuning process, ensuring that the adjustments to the model are precisely tailored to enhance performance on the specified task.

As for DistilmBERT, the maximum length of input sequences was set to 512. The AdamW optimizer was employed for training with a learning rate set to $1.0e-4$ and a batch size of 20. This model was trained for 5 epochs.

\subsection{Results and Discussions}

\begin{table*}[htb!]
    \centering
    \scalebox{.9}{
    \begin{tabular}{lllllllll}
    \toprule
    \multicolumn{1}{c}{\multirow{2}{*}{Method}} & \multicolumn{2}{c}{Subtask A - Monolingual} & & \multicolumn{2}{c}{Subtask A - Multilingual} & & \multicolumn{2}{c}{Subtask B} \\
    \cmidrule{2-3} \cmidrule{5-6} \cmidrule{8-9}
    \multicolumn{1}{c}{} & \multicolumn{1}{c}{Dev} & \multicolumn{1}{c}{Test} & & \multicolumn{1}{c}{Dev} & \multicolumn{1}{c}{Test} & & \multicolumn{1}{c}{Dev} & \multicolumn{1}{c}{Test} \\ \midrule
    LR & 0.673 &\multicolumn{1}{c}{Test results will be released in Feb} & & 0.473 &\multicolumn{1}{c}{-} & & 0.251 &\multicolumn{1}{c}{-} \\
    MultinomialNB & 0.555 &\multicolumn{1}{c}{-} & & 0.483 &\multicolumn{1}{c}{-} & & 0.435 &\multicolumn{1}{c}{-} \\
    XGBoost & 0.692 &\multicolumn{1}{c}{-} & & 0.515 &\multicolumn{1}{c}{-} & & 0.540 &\multicolumn{1}{c}{-} \\ \midrule
    XLM-RoBERTa & 0.783 &\multicolumn{1}{c}{-} & & 0.679 &\multicolumn{1}{c}{-} & & 0.735 &\multicolumn{1}{c}{-} \\
    LoRA-RoBERTa & 0.783 &\multicolumn{1}{c}{-}  & & 0.726 &\multicolumn{1}{c}{-} & & 0.735 &\multicolumn{1}{c}{-} \\
    Majority voting & 0.735 &\multicolumn{1}{c}{-} && 0.728 &\multicolumn{1}{c}{-} & & 0.717 &\multicolumn{1}{c}{-} \\
    DistilmBERT & 0.702 &  &  &  0.670 &   & & 0.629 &\multicolumn{1}{c}{-} \\
    \bottomrule
    \end{tabular}}
    \caption{Comparative performance of machine learning and transformer-based models on text classification subtasks}
    \label{tab:results}
\end{table*}

In our experiments, we evaluated various models on three distinct subtasks: Subtask A - Monolingual, Subtask A - Multilingual, and Subtask B. Each subtask involved both development (Dev) and test phases, although test results are not provided for some models. The models tested included traditional machine learning algorithms - LR, MultinomialNB, and XGBoost - as well as advanced transformer-based models like XLM-RoBERTa, LoRA-RoBERTa, and DistilmBERT. Additionally, we employed a majority voting ensemble method combining XLM-RoBERTa and LoRA-RoBERTa.

Analyzing the results given in Table \ref{tab:results}, it is evident that transformer-based models, particularly LoRA-RoBERTa, outperformed traditional machine learning models across all subtasks. For instance, in Subtask A - Monolingual and Subtask B, LoRA-RoBERTa achieved the highest Dev scores. Notably, DistilmBERT, while trailing behind its counterparts, significantly higher than the scores of classical models. Moreover, the majority voting ensemble method further showed its effectiveness, especially in the multilingual subtask, by achieving high scores and demonstrating its capability in integrating different model strengths.

\subsection{Subtask A - Monolingual}

In the monolingual Subtask A, Transformer models, particularly LoRA-RoBERTa and XLM-RoBERTa, achieved the highest Dev scores of 0.783, significantly outperforming traditional models. DistilmBERT also demonstrated superior performance over traditional models but performed less effectively than other fine-tuned LLMs. Traditional models like LR and MultinomialNB showed weaker performance, with XGBoost being the strongest among them with a score of 0.692. This disparity suggests the traditional ML models' limited ability to capture contextual relationships in text, unlike Transformer models.

\subsection{Subtask A - Multilingual}

In the challenging multilingual Subtask A, the majority voting ensemble method outperformed individual models, achieving the highest Dev score of 0.728. This surpassed LoRA-RoBERTa's 0.726, XLM-RoBERTa's 0.679, and DistilmBERT's 0.670. The ensemble's success highlights its capability to effectively integrate different model strengths, proving particularly adept at handling the complexities of multilingual data.

\subsection{Subtask B}

In Subtask B, both LoRA-RoBERTa and XLM-RoBERTa maintained their lead with identical Dev scores of 0.735. The majority voting ensemble also showed strong performance, scoring 0.717. DistilmBERT performed worse among LLMs but outperformed the ML models. Traditional models, however, were significantly outperformed, with XGBoost again being the best among them with a score of 0.540.

\section{Conclusions}

The results across all subtasks consistently demonstrate the exceptional performance of Transformer models, particularly LoRA-RoBERTa, in text classification tasks. Their advanced capability in natural language understanding and processing, along with their proficiency in capturing subtle nuances, make them superior to traditional models. DistilmBERT, representing a more streamlined transformer approach, also outperforms traditional ML models, highlighting the evolving landscape of NLP tools. Additionally, ensemble methods, like majority voting, have proven effective, suggesting that a hybrid strategy combining various model types could be advantageous in certain cases.

\bibliographystyle{IEEEtran}
\bibliography{custom}

\end{document}